\documentclass{article}

\usepackage{microtype}
\usepackage{graphicx}
\usepackage{subcaption}
\usepackage{booktabs} 

\usepackage{hyperref}

\usepackage[accepted]{icml2026}

\usepackage{amsmath}
\usepackage{amssymb}
\usepackage{mathtools}
\usepackage{amsthm}

\usepackage[capitalize,noabbrev]{cleveref}
\usepackage{xcolor}

\theoremstyle{plain}

\theoremstyle{definition}

\theoremstyle{remark}

\usepackage[disable,textsize=tiny]{todonotes}

\icmltitlerunning{Latent Cache Flow}

\begin{document}

\twocolumn[
\icmltitle{Latent Cache Flow: Model-to-Model Communication Without Text}

\icmlsetsymbol{equal}{*}

\begin{icmlauthorlist}
\icmlauthor{Maximillian Rossi}{columbia,equal}
\icmlauthor{Prajwal Raghunath}{columbia,equal}
\icmlauthor{Eugene Wu}{columbia}
\end{icmlauthorlist}

\icmlaffiliation{columbia}{Columbia University}

\icmlcorrespondingauthor{Maximillian Rossi}{mfr2178@columbia.edu}




\icmlkeywords{Latent Cache Flow, LCF, Cache-to-Cache, C2C, KV Cache, Cache Communication, Cross-Model Communication, Cross-Context Communication, Multi-Model Systems, LLM Communication, LLM Inference, Parameter Efficiency, Adapter Efficiency, Bottleneck Compression, Key-Value Compression, Multi-Head Latent Attention, Text-to-Text Communication}

\vskip 0.3in
]


\printAffiliationsAndNotice{\icmlEqualContribution\ Latent Cache Flow is patent pending. U.S. Provisional Patent Application No. 64/065,974.}
\begin{abstract}
LLM agents today communicate via text, which incurs considerable latency and information loss due to the need to autoregressively decode the sharer model's state and encode at the receiver model. Recent work such as Cache-to-Cache \citep[C2C;][]{fu2026cache} seeks to exchange KV caches by learning adapters that translate sharer KV matrices to the receiver model. However, the adapters are large and expensive to train, and translate individual tokens, which requires the target context to be identical. This is unsuitable for agent communication, where the LLMs have differing context.

We introduce Latent Cache Flow (LCF). To address efficiency, we observe that keys and values can be jointly translated and compressed, reducing the adapter to about 4\% of C2C's size. To address differing context, we design the adapter to transmit a summary of new information that the target model does not have. Our early experiments show that a pruned 13~\text{MB} LCF adapter can be more accurate than C2C at 956~\text{MB} in shared-context settings; for different contexts, LCF improves F1 by 7.5\% and Exact Match by 23\% while $8.5\times$ faster than text-based communication.

\end{abstract}
\section{Introduction}
\label{sec:intro}

Language-model workflows involve multiple models with different roles, tools, and contexts \citep{yao2023react, du2023debate, wu2023autogen, chen2024agentverse}. These models must communicate to combine their work, and they usually do so through generated text. For example, one model may summarize a document chunk, and another may use that summary to answer a question over a different chunk. This text-mediated channel is slow because the receiver must wait for the sender to decode, and lossy because much of the sender's internal state is compressed into discrete tokens.

Cache-to-Cache (C2C) \citep{fu2026cache, dery2026latent} addresses the text bottleneck by exchanging KV caches directly. In C2C, the sharer and receiver process the same input, and a learned adapter fuses their caches at matching token positions. This bypasses intermediate text generation, but it makes C2C position-wise by design. The assumption is restrictive for deployed multi-model systems, where models are used because they handle different sub-tasks, tools, or context. When their contexts differ, their token positions no longer align, and position-wise cache exchange does not apply. C2C also introduces substantial adapter overhead: for a 1.1B-parameter sharer--receiver pair, its adapter is 956~\text{MB}.

We introduce Latent Cache Flow (LCF) to address both limitations. LCF treats cache transfer as compressed communication rather than token-by-token translation. It replaces separate key and value fusers with a shared low-dimensional cache channel, and uses layer pruning to remove receiver layers that do not benefit from cache updates. We extend LCF to cross-context communication with LCF-X. Instead of translating each sharer token into a receiver-aligned cache update, LCF-X summarizes the sharer's full KV cache into a fixed-size tensor that the receiver can condition on.

\begin{figure}[!htbp]
\centering
\includegraphics[width=1\linewidth]{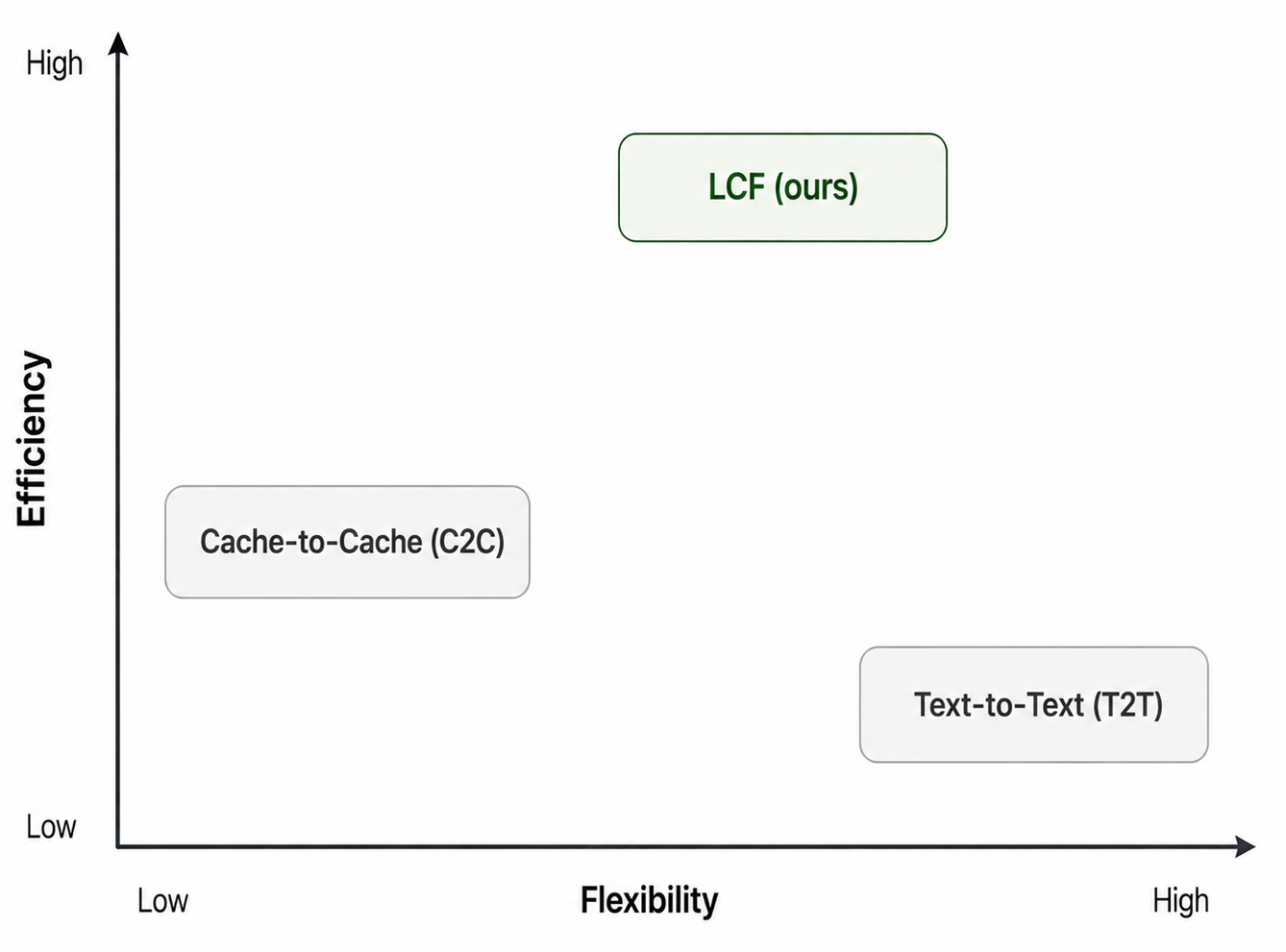}
\caption{\footnotesize Conceptual efficiency--flexibility trade-off. LCF improves prior cache-level transfer efficiency while relaxing C2C's alignment constraint toward T2T-like flexibility.}
\label{fig:efficiency_flexibility}
\end{figure}

Text-to-text (T2T) and C2C represent opposite extremes. T2T is flexible because the sharer can send free-form text, but it pays decoding latency and token-projection loss. C2C avoids decoding by exchanging caches at matching positions, but requires aligned inputs. Figure~\ref{fig:efficiency_flexibility} illustrates the middle ground targeted by LCF and LCF-X: compressed cache-level communication that improves efficiency while relaxing the shared-context constraint.

\section{Background: Cache-to-Cache}

\label{sec:c2c_to_lcf}
The natural target for cross-model transfer is the KV cache. This cache is the residue of the sharer’s forward pass and holds the per-layer key and value tensors that summarize its reading of the prompt. However, direct cache reuse is not feasible across different architectures. The sharer and receiver may differ in the number of layers, attention heads, head dimensions, or the geometry of the learned latent space. As a result, the receiver cannot consume the sharer’s KV tensors verbatim. Cache-to-Cache (C2C) addresses this challenge by treating the sharer’s cache as a learned semantic conditioning signal. C2C trains a fuser module that takes both models’ caches as input and generates a residual update, which is added to the receiver cache before decoding.

\begin{figure}[!htbp]
\centering
\includegraphics[width=1\linewidth]{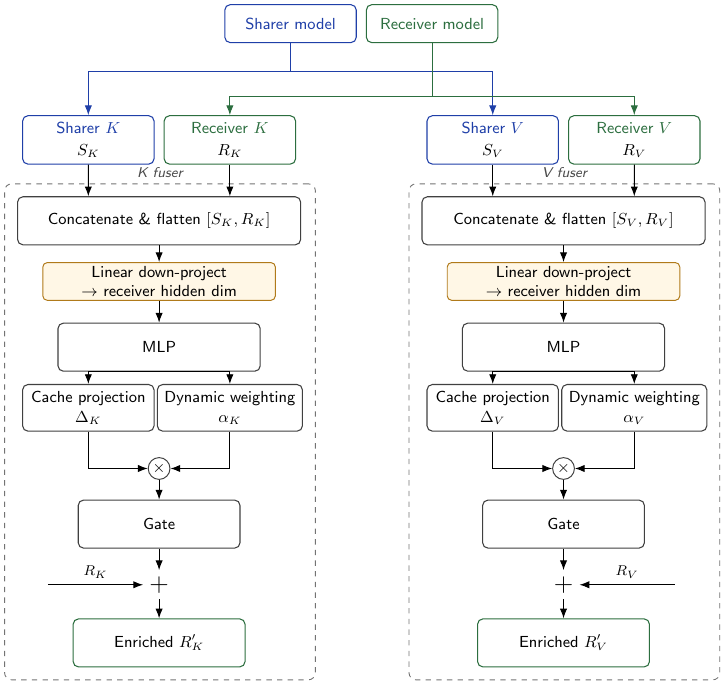}
\caption{\footnotesize C2C pipeline. Yellow denotes projection and dimension-changing operations; blue denotes sharer outputs, and green denotes receiver states.}
\label{fig:c2c_architecture}
\end{figure}

Figure~\ref{fig:c2c_architecture} illustrates the C2C fuser, which contains two independent duplicate modules for the key and value tensors. We describe the key side; values are handled similarly.

After the sharer and receiver process the prompt, they emit layer-aligned multi-head keys $S_K$ and $R_K$. The fuser concatenates these keys and flattens them along the head dimension to form a joint per-token feature vector. A linear layer then projects this vector down to the receiver’s hidden dimension. An MLP performs feature fusion, mixing the sharer and receiver projections into a joint encoding.

The fused features split into two parallel paths. The projection path produces a per-head cache update $\Delta_K$ in the receiver’s $(H_t, D_h)$ shape. The dynamic weighting path produces a per-head scalar $\alpha_K$ that re-weights the update. Their element-wise product is masked by a learned Gumbel-sigmoid gate, a differentiable relaxation of an on/off decision that is annealed toward a binary layer-selection at inference. Finally, the result is added to $R_K$ as a residual, yielding the enriched cache $R'_K$.

\section{Latent Cache Flow}
\label{sec:architecture}

In this section, we first describe how the design of Latent Cache Flow (LCF) improves the efficiency of KV transfer when the contexts match.  We then describe an extension called LCF-X that pre-processes the sharer model's KV cache to support communication between differing contexts.

\subsection{Latent Cache Flow for Exact Context Sharing}

We address two main sources of inefficiencies in C2C.   
First, C2C processes keys and values independently by duplicating the fuser pipeline for each. However, keys and values are projections of the same hidden state. Recent architectures, such as Gemma 4 \citep{gemma4modelcard}, showcase this redundancy through unified key-value tensors for global attention layers. LCF replaces C2C’s dual pipelines with a single shared key-value pipeline.

Second, C2C keeps communication in the receiver's high-dimensional cache space. It concatenates the sharer and receiver caches, fuses them at full width, and outputs receiver-shaped residuals. This makes transfer expensive: each cache type needs its own full-width fuser.

LCF instead transfers information through a low-dimensional latent bottleneck. This bottleneck is motivated by two observations: KV states can be compressed within a model \citep{deepseek2024v2}, and cross-model representations can be approximately linearly aligned 
\citep{chen2025linear, dery2026latent}. LCF combines these observations by projecting the joint key--value representation into a compact latent space, mixing it, and up-projecting the result back into receiver-cache residuals.

\begin{figure}[t]
\centering
\includegraphics[width=1\linewidth]{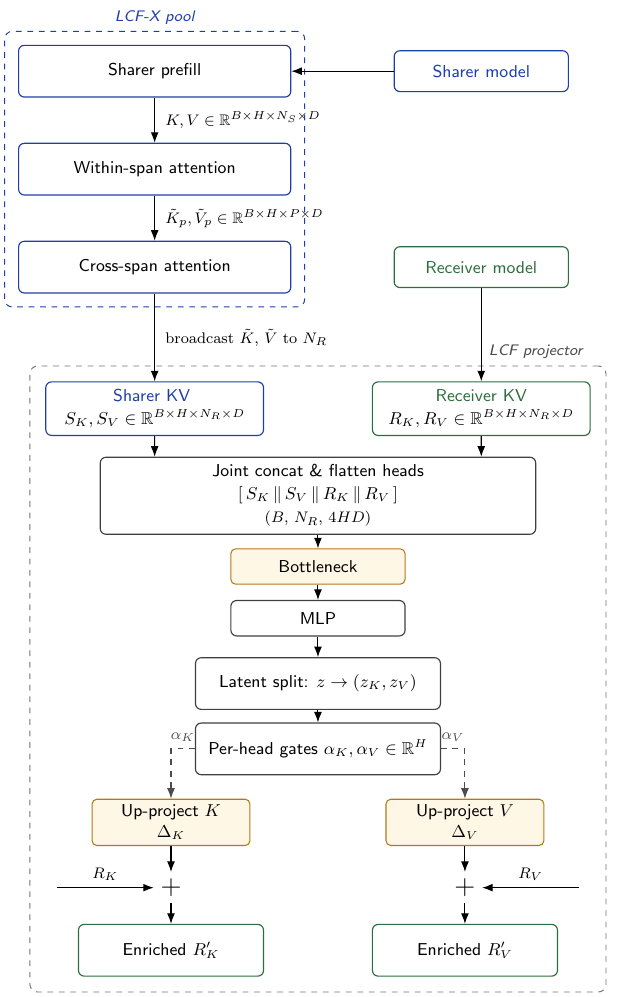}
\caption{\footnotesize LCF/LCF-X architecture. Blue denotes sharer-only states, green denotes receiver-only states, and yellow denotes projection/compression operations. Sharer-side pooling enables cross-context summaries; the shared LCF projector compresses KV inputs and up-projects receiver-cache residuals.}
\label{fig:LCF_architecture}
\end{figure}

Figure~\ref{fig:LCF_architecture} shows the full architecture in two parts: the LCF-X pooling module at the top and the core LCF projector at the bottom. This section focuses on the core LCF projector, which assumes token-aligned sharer and receiver caches.

\textbf{1. Input.} Each receiver transformer layer receives a pair of sharer cache $(\mathbf{S}_K, \mathbf{S}_V)$ and receiver cache $(\mathbf{R}_K, \mathbf{R}_V)$.

\textbf{2. Concat \& Flatten} The KV cache tensors from the Sharer and Receiver are concatenated and flattened along the head dimension into a single tensor.

\textbf{3. Shared Latent Pipeline.} The concatenated tensor $\mathbf{x}$ is linearly down-projected to $d$ dimensions: $\mathbf{z} = W_{\mathrm{down}}\mathbf{x}$, followed by a two-layer MLP. We select $d \in \{64, 128, 256, 512\}$ as a hyperparameter based on downstream accuracy (Figure ~\ref{fig:pareto}).

\textbf{4. Latent split.} The latent tensor is then split into separate key and value tensors: 
$[\mathbf{z}_K, \mathbf{z}_V] = \mathbf{z}$ where 
$\mathbf{z}_i \in \mathbb{R}^{B \times N \times d/2}$. At this point, 
the keys and values will be reconstructed independently. 

\textbf{5. Gates.} Per-head gates $\alpha_i \in \mathbb{R}^{h_r}$ are produced from the full latent $\mathbf{z}$, allowing K and V updates to be independently scaled using shared latent features.

\textbf{6. Upward project.} The Key and Value tensors are projected back to receiver-cache size:
\[
\Delta_i = W_{\mathrm{up}}^i \mathbf{z}_i, \quad i \in \{K, V\}.
\]
We adopt C2C's per-layer Gumbel-sigmoid gate, producing $g_i \in \{0, 1\}$ at inference. The receiver cache is updated as:
\[
\mathbf{R}'_i = \mathbf{R}_i + g_i \cdot (\alpha_i \odot \Delta_i), \quad i \in \{K, V\}.
\]

\textbf{7. Enriched receiver KV.} The gated residuals update the receiver's K and V caches.

\textbf{Adapter Parameters.} For a pair of 1.1B parameter models, C2C requires a 477.8M-parameter adapter model.   At $d = 128$, LCF only requires 19.4M parameters across 28 layers (693K per layer)---$24.6\times$ fewer than C2C.

\subsection{LCF-X: Cross-Context Communication}

\label{sec:lcfx}
LCF reduces C2C's adapter overhead, but it still assumes shared context: the sharer and receiver must process the same prompt so that each receiver token has a matching sharer token. This position-wise requirement prevents cross-context communication. LCF-X removes it by replacing matched sharer tokens with a position-free summary of what the sharer learned from its own context. As shown in Figure~\ref{fig:LCF_architecture}, LCF-X adds a sharer-side pooling module before the standard LCF projector; the receiver therefore conditions on the sharer's learned context, not on a token-aligned variation of its own prompt.

We divide the sharer's context into $P$ spans $S_1,\ldots,S_P$. A span is a contiguous block of source tokens, such as a paragraph, passage, or fixed-size chunk. The first part of Figure~\ref{fig:LCF_architecture} shows this pooling step. LCF-X uses attention as a pooling operation: within each span, a learned query attends only to the tokens in that span and produces one key--value summary. Across all spans, this gives
\[
\tilde{\mathbf{K}}, \tilde{\mathbf{V}} \in \mathbb{R}^{B \times H \times P \times D}.
\]
The $P$ summaries keep span identity before global pooling. Larger $P$ gives cross-span attention more local summaries to weight; smaller $P$ gives it fewer, coarser summaries. Because the same pooling query is reused for every span, $P$ can vary at inference; one checkpoint can summarize one passage, five passages, or a sliding set of chunks.
Appendix~\ref{app:lcfx_span_invariance} gives the span-invariance analysis.
    
LCF-X then pools across spans. A second attention step aggregates the $P$ span summaries into one per-head sharer summary,
\[
\mathbf{K}^*, \mathbf{V}^* \in \mathbb{R}^{B \times H \times 1 \times D}.
\]
This summary is broadcast across the receiver sequence and replaces the sharer's matched-position cache in LCF. The lower part of Figure~\ref{fig:LCF_architecture} shows that, after this replacement, the standard LCF projector and receiver cache are unchanged.

LCF and LCF-X address different limits of C2C. LCF replaces C2C's high-dimensional, cache-specific fusers with a compressed shared key--value pipeline. LCF-X removes the shared-context assumption by replacing matched-position sharer caches with pooled, position-free summaries. The next section tests whether these changes preserve cache-transfer accuracy while reducing adapter overhead.

\section{Experiments}
\label{sec:experiments}
We first evaluate LCF as compared to C2C in a shared-context setting, and then evaluate LCF-X in a cross-context setting as compared to T2T communication, where models observe different contexts and must share novel information.

\subsection{Shared-Context}
\label{sec:shared_context}

\subsubsection{Experimental Setup}

The shared-context setting assumes aligned token positions between models. It tests LCF's compression claim: whether a smaller latent fuser can match or improve C2C's full-width cache fusion.

The sharer is Qwen2.5-0.5B-Instruct \citep{qwen25blog}, and the receiver is Qwen3-0.6B \citep{qwen3technicalreport}. We use OpenHermes 2.5 \citep{openhermes2.5}, matching C2C's training data, and keep both base models frozen. For C2C, we use the released full-schedule \texttt{nics-efc/C2C\_Fuser} checkpoint. We train LCF under the same public C2C recipe, changing only the fuser architecture. Full hyperparameter, filtering, and compute details are provided in Appendix~\ref{app:training}.

We evaluate LCF's sensitivity to bottleneck dimension with $d=\{64,128,256,512\}$. We also evaluate layer pruning, which removes receiver layers whose learned gates are near zero. This tests whether LCF can further reduce adapter size without losing downstream accuracy.

We evaluate on four logit-based zero-shot multiple-choice benchmarks. We use MMLU-Redux (MMLU-R) \citep{gema2024mmlu}, ARC-Challenge (ARC-C) \citep{allenaiarc}, and OpenBookQA (OBQA) \citep{OpenBookQA2018} from C2C's benchmark suite, and replace C-Eval with MMLU-Pro \citep{wang2024mmlupro} for stronger English-language coverage. Logit-based scoring avoids free-form parsing noise by reading the predicted answer directly from the model's logits over the answer choices. We also compare with an optimal-routing frontier (ORF), a per-question oracle that selects the correct answer when either base model is correct.

\subsubsection{Training Cost}
\label{sec:adapter_training}

Under the matched pipeline, training time is dominated by frozen model prefills, not adapter computation. Each optimizer step runs a sharer prefill, a receiver prefill, cache fusion, and a backward pass through the adapter. This approach holds the communication workload fixed for both adapters, rather than optimizing around LCF.

With this setup, LCF-128 trains for 300 optimizer steps in about 4.5 hours on a single Colab A100. Each optimizer step takes 52 seconds, with an effective batch size of 260. For comparison, our C2C run on a single A100 takes 66 seconds per optimizer step after warmup, measured with an effective batch size of 256. We use 300 LCF steps because both the published C2C curve and our C2C reproduction achieve their best MCQ accuracy at step 250. Later checkpoints do not improve downstream accuracy.

The modest per-step wall-clock difference reflects where training time is spent. Most computation comes from the frozen sharer and receiver forward passes, which both methods share. The smaller LCF adapter reduces trainable parameters, adapter memory, and adapter-side backward cost. However, these savings are partly hidden by the frozen-model prefill cost. When KV caches are precomputed or reused, adapter-side costs become more visible. In those settings, smaller adapters may provide larger practical savings.

\subsubsection{Shared-Context Results}

\begin{figure}[!htbp]
\centering
\includegraphics[width=\linewidth]{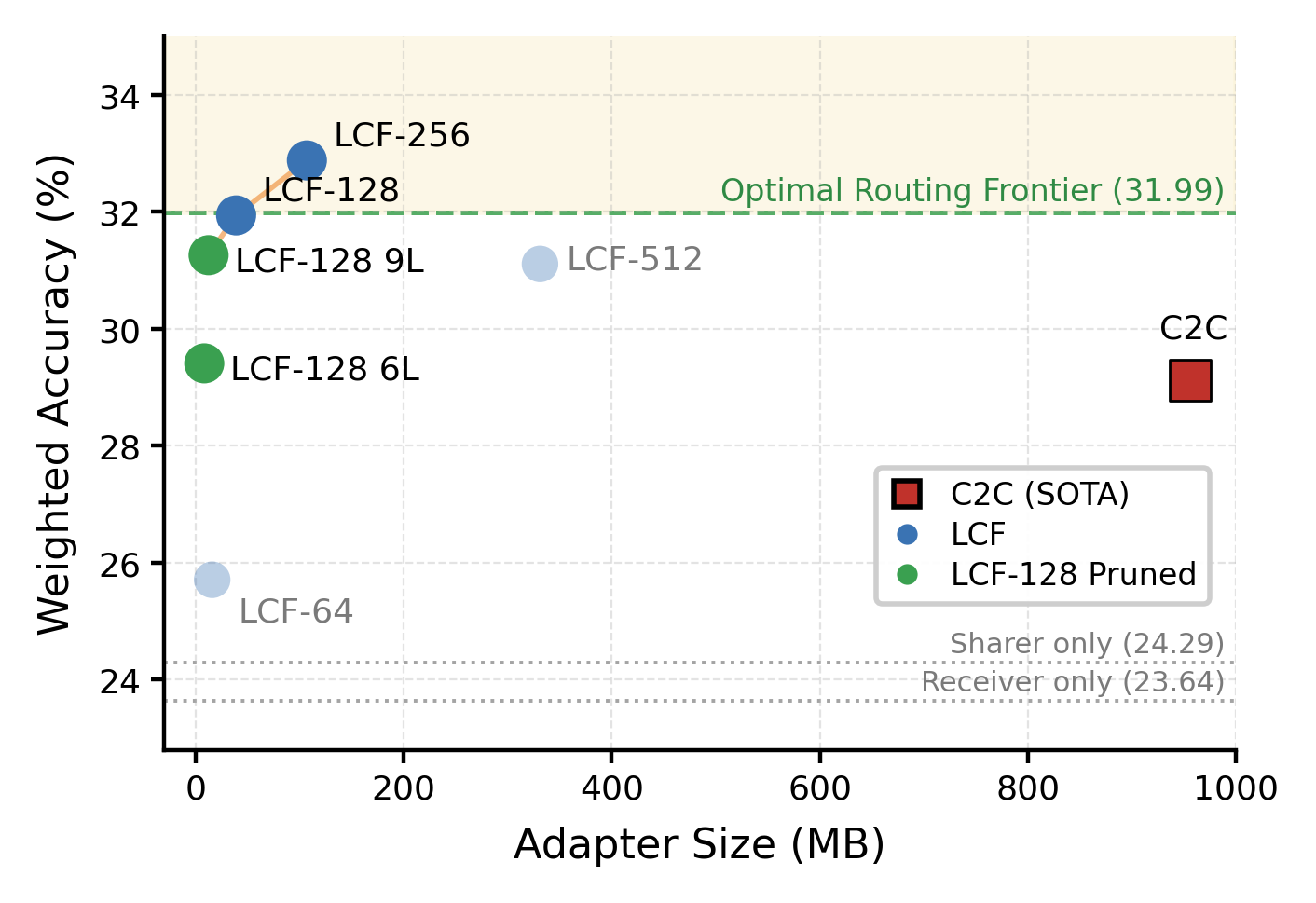}
\caption{Accuracy--efficiency frontier for shared-context communication. Weighted accuracy versus bf16 projector footprint. 9L denotes the 9-layer pruned variant.}
\label{fig:pareto}
\end{figure}

Figure~\ref{fig:pareto} summarizes the shared-context results, including both bottleneck scaling and layer-pruned variants. LCF establishes an accuracy--efficiency frontier for inter-model communication, outperforming C2C at substantially lower adapter overhead. The frontier is traced by the progression from LCF-128-9L to LCF-128, and finally to LCF-256. Increasing latent capacity improves accuracy while keeping adapter size much smaller than C2C.

LCF-256 achieves the strongest overall performance, reaching a weighted average of 32.88 across the evaluated benchmarks while using only $107$ MB. LCF-128 provides a particularly efficient operating point near the Pareto knee, achieving 31.94 weighted accuracy with only 39 MB of overhead. Aggressively pruned variants remain competitive, the 9-layer configuration reaches 31.26 weighted accuracy using only 13~\text{MB}.

The frontier also reveals both under- and over-capacity regimes. LCF-64 underperforms, suggesting insufficient communication bandwidth. In contrast, LCF-512 increases adapter size substantially but provides weaker accuracy than LCF-256. This non-monotonic behavior suggests that the bottleneck acts as a compression mechanism and as a regularizer for communication.

Notably, LCF-256 exceeds the optimal routing frontier (ORF) on ARC-C and MMLU-Pro, suggesting gains that retrospective routing cannot capture, whether through transfer or shifted decision boundaries.

\begin{table}[!htbp]
\centering
\footnotesize
\resizebox{\columnwidth}{!}{%
\begin{tabular}{lcccc|c}
Method & MMLU-R & ARC-C & OBQA & MMLU-Pro & Wtd avg \\
\midrule
Receiver          & 34.91 & 39.91 & 39.20 & 16.17 & 23.64 \\
C2C               & 42.54 & 54.00 & 50.80 & 19.57 & 29.13 \\
\textbf{LCF-128}  & 45.17$^{*}$ & 56.09 & \textbf{52.60} & 22.57$^{*}$ & 31.94 \\
\textbf{LCF-256}  & \textbf{47.53}$^{*}$ & \textbf{57.65}$^{*}$ & 51.80 & \textbf{22.87}$^{*}$ & \textbf{32.88} \\
ORF               & 48.90 & 54.43 & 55.40 & 20.95 & 31.99 \\
\end{tabular}%
}
\caption{Zero-shot logits accuracy (\%). Wtd avg uses test sizes 5632/1150/500/12032. $^{*}$ $p<0.05$ vs.\ C2C.}
\label{tab:main_LCF}
\end{table}

\subsubsection{Layer Pruning}
\label{sec:layer_pruning}

Layer pruning reveals that LCF communication concentrates in a small subset of receiver layers. We prune LCF-128 by removing layers whose learned Gumbel-sigmoid gates converge toward zero during training.

\begin{table}[!htbp]
\centering
\footnotesize
\setlength{\tabcolsep}{6pt}
\renewcommand{\arraystretch}{1.0}
\begin{tabular}{lccc}
\toprule
Config & Wtd Avg & Size (MB) & vs.\ C2C \\
\midrule
C2C (reference) & 29.13 & 956 & $1\times$ \\
\midrule
LCF-128 full (28L)    & \textbf{31.94} & 39 & $24.5\times \downarrow$ \\
LCF-128 pruned (19L)  & 31.92 & 26 & $36\times \downarrow$ \\
LCF-128 critical (9L) & 31.26 & 13 & $76\times \downarrow$ \\
LCF-128 critical (6L) & 29.41 & \textbf{8} & \textbf{$115\times \downarrow$} \\
\bottomrule
\end{tabular}
\caption{Layer pruning of LCF-128 with C2C as reference. Parentheses denote retained layers. Weighted average is computed across MMLU-R, ARC-C, OBQA, and MMLU-Pro. Size reports bf16 projector footprint only; ``vs.\ C2C'' shows the reduction in projector size relative to C2C.}
\label{tab:pruning}
\end{table}

As shown in Table~\ref{tab:pruning}, performance degrades gradually as layers are removed, indicating that LCF communication is highly concentrated rather than uniformly distributed across the network. The 19-layer pruned model reduces projector size from 39 MB to 26 MB with almost no loss in weighted accuracy. The 9-layer critical subset continues to outperform C2C while requiring only 13 MB of overhead.

The aggressive 6-layer variant approximately matches C2C performance (29.41 vs.\ 29.13) using only 8 MB of projector parameters, corresponding to a $115\times$ reduction in overhead. This suggests that a small subset of receiver layers accounts for most inter-model communication gains. Layer indices and pruning details are provided in Appendix~\ref{app:pruning}.

\subsection{Cross-Context Information Transfer}
\label{sec:cross_context}
\subsubsection{Experimental Setup}

The cross-context setting tests whether LCF-X can perform cache-level transfer without being restricted to token-aligned sharer and receiver inputs.

We create the cross-context setting by partitioning HotpotQA's distractor context \citep{yang2018hotpotqa}. In the standard distractor setting, each question is paired with ten paragraphs: two supporting paragraphs and eight distractors. We split these ten paragraphs into two disjoint sets of five, with each set containing one supporting paragraph and four distractors. One set is assigned to the sharer and the other to the receiver. This partition gives the models different contexts while preserving the original two-hop evidence structure. Since the receiver sees only one supporting paragraph, exact-match answering generally requires new context from the sharer's partition.

Both models are Qwen3-0.6B instances, isolating communication effects from model capacity. We first establish three receiver-only baselines to measure how parametric knowledge and evidence availability affect performance: (i) question-only, where the receiver answers using internal knowledge without supporting paragraphs, (ii) half-context, where the receiver sees only its partition, and (iii) full-context, where the receiver sees all ten paragraphs.

We then compare T2T with LCF-X. In the T2T baseline, the sharer reads its partition and generates a text message, which is appended to the receiver's input. We evaluate T2T with communication budgets of 50, 100, 150, and 200 generated tokens. In LCF-X, the sharer's KV cache is pooled into a fixed-size cache summary and passed through the LCF projector, enabling cross-context cache transfer without token alignment.

We train LCF-X directly on the partitioned HotpotQA training set. Both the sharer and receiver are frozen; only the LCF projector and hierarchical pooling queries are trainable. Training runs for one epoch, 276 optimizer steps, with an effective batch size of 256. The process takes roughly 75 minutes on a single NVIDIA RTX PRO 6000 Blackwell GPU. Full training details are provided in Appendix~\ref{app:lcfx_training_details}.

We evaluate answer quality on the HotpotQA-bridge ($n = 5{,}899$) evaluation split. We report Exact Match (EM), which measures whether the predicted answer matches the ground truth string exactly. We also report token-level F1, which measures partial overlap between predicted and reference answers. We additionally report latency in terms of time-to-first-token (TTFT) and time-to-end-of-answer (TTEoA).

\subsubsection{Cross-Context Results}

LCF-X improves both quality and latency in the tested cross-context setting. On HotpotQA-bridge, LCF-X reaches 35.13 F1 and 25.28 EM, outperforming the strongest T2T baseline, T2T Max 200, by +2.47 F1 and +4.75 EM. These gains are statistically robust under both paired F1 and exact-match tests (Wilcoxon $p = 5.42 \times 10^{-5}$; McNemar $p = 4.23 \times 10^{-12}$).

\begin{table}[!htbp]
\centering
\setlength{\tabcolsep}{7pt}
\begin{tabular}{lrrrr}
\toprule
Protocol & F1 & EM & TTFT & TTEoA \\
         &    &    & (ms) & (ms)  \\
\midrule
\multicolumn{5}{l}{\textbf{Receiver}} \\
Question-only & 7.48 & 1.37 & 11 & 129 \\
Half context  & 24.62 & 14.48 & 14 & 104 \\
Full context  & 29.98 & 16.77 & 18 & 117 \\
\midrule
\multicolumn{5}{l}{\textbf{T2T}} \\
Max 50  & 30.81 & 18.83 & 250 & 342 \\
Max 100 & 31.68 & 19.75 & 290 & 381 \\
Max 150 & 32.58 & 20.46 & 327 & 419 \\
Max 200 & 32.66 & 20.53 & 410 & 502 \\
\midrule
\multicolumn{5}{l}{\textbf{LCF}} \\
LCF-X & \textbf{35.13} & \textbf{25.28} & \textbf{48} & \textbf{94} \\
\bottomrule
\end{tabular}
\caption{Cross-context results on HotpotQA-bridge ($n{=}5{,}899$). TTFT is time to first token; TTEoA is time to end of answer.}
\label{tab:final_results}
\end{table}

The receiver-only baselines show that evidence availability matters. Moving from question-only to half-context raises F1 from 7.48 to 24.62. Providing full context raises it further to 29.98. Both T2T and LCF-X exceed the full-context receiver baseline. This suggests that communication recovers missing paragraphs and changes how evidence reaches the receiver. Information now arrives through the sharer's forward pass, either as generated text or as a pooled cache summary.

The key comparison is therefore between communication methods. Increasing the T2T budget from 50 to 200 tokens improves F1 from 30.81 to 32.66, but increases TTFT from 250 ms to 410 ms. T2T buys modest quality gains with additional decoding. LCF-X avoids this trade-off: it reaches higher quality than every T2T budget while reducing TTFT to 48 ms and TTEoA to 94 ms. Because TTEoA also depends on the generated answer length, we treat TTFT as a cleaner measure of communication overhead and report TTEoA as the end-to-end response time.

\begin{figure}[!htbp]
\centering
\includegraphics[width=1\linewidth]{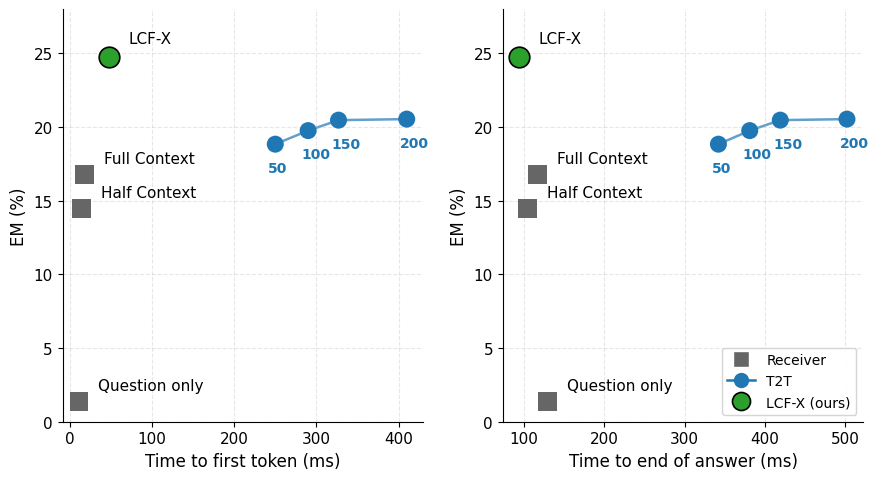}
\caption{\footnotesize Quality--latency trade-off on HotpotQA-bridge.}
\label{fig:hotpotqa_latency_quality}
\end{figure}

Figure~\ref{fig:hotpotqa_latency_quality} summarizes the quality--latency frontier. LCF-X lies above and to the left of the T2T sweep, indicating higher answer quality at lower communication latency.

\section{Limitations and Future Work}
While these results are promising, they are an initial validation of LCF. Our shared-context experiments use one model pair, Qwen2.5-0.5B-Instruct to Qwen3-0.6B, and our cross-context experiments use two Qwen3-0.6B instances. Although C2C works across model families and larger scales \citep{fu2026cache}, LCF testing is future work.

Our different context setting is also deliberately controlled. Partitioned HotpotQA assesses if information can move through the channel, though we cannot fully separate transfer from tuned adaptation of the receiver. Richer agentic workflows with longer contexts, tool outputs, heterogeneous roles, and multi-turn communication remain future work.

\section{Conclusion}
\label{sec:conclusion}

Multi-model language workflows need communication channels that are fast, lightweight, and compatible with distinct model contexts. Text-to-text communication is flexible but sequential and lossy. Prior cache-based methods avoid text generation, but they require shared, token-aligned inputs and large adapters for each model pair.

Latent Cache Flow (LCF) shows that cache communication does not need full-width cache transfer. In the shared-context setting, LCF matches or outperforms C2C across all four benchmarks with a $13~\mathrm{MB}$ adapter, compared with C2C's $956~\mathrm{MB}$ adapter. This result shows that cache information can flow from one model to another through a latent channel.

LCF-X then moves cache communication toward greater context flexibility. On partitioned HotpotQA, LCF-X transfers information across distinct contexts, improves exact-match accuracy by $23\%$ over text communication, and reduces time-to-first-token by $8.5\times$. Together, these results move inter-model communication toward a different interface: cache-level rather than textual, compressed rather than full-width, and flexible beyond token-aligned contexts.


\bibliography{references}
\bibliographystyle{icml2026}

\newpage
\appendix
\onecolumn
\section{Full Results Overview}
\label{app:full_results}

Table~\ref{tab:app_full_overview} lists all methods evaluated in this paper. T2T is a cited baseline from C2C. We do not re-evaluate it. ORF is the optimal-routing frontier. This is a per-question oracle that picks whichever base model gives the correct answer. ORF can therefore be seen as a retrospective optimal routing system.

\begin{table}[htbp]
\centering
\footnotesize
\setlength{\tabcolsep}{3pt}
\begin{tabular}{lccccccc}
\toprule
Method & MMLU-R & ARC-C & OBQA & MMLU-Pro & Wtd Avg & Size (MB) & Params/layer \\
\midrule
Receiver only & 34.91 & 39.91 & 39.20 & 16.17 & 23.64 & 0 & 0 \\
Sharer only & 38.65 & 41.30 & 44.80 & 15.09 & 24.29 & 0 & 0 \\
T2T$^{\dagger}$ & 41.03 & 49.48 & 44.00 & -- & -- & -- & -- \\
C2C & 42.54 & 54.00 & 50.80 & 19.57 & 29.13 & 956 & 17.1M \\
LCF-64 & 38.32 & 46.17 & 43.00 & 17.14 & 25.71 & 16 & 0.28M \\
LCF-128 & 45.17$^{***}$ & 56.09 & 52.60 & 22.57$^{***}$ & 31.94 & 39 & 0.69M \\
\textbf{LCF-256} & \textbf{47.53}$^{***}$ & \textbf{57.65}$^{**}$ & 51.80 & \textbf{22.87}$^{***}$ & \textbf{32.88} & 107 & 1.9M \\
LCF-512 & 45.47$^{***}$ & 56.61$^{*}$ & \textbf{54.40}$^{*}$ & 20.98$^{***}$ & 31.11 & 331 & 5.9M \\
LCF-128-9L$^{\ddagger}$ & 45.65$^{***}$ & 54.70 & 51.40 & 21.45$^{***}$ & 31.26 & 13 & 0.69M \\
LCF-128-6L$^{\ddagger}$ & 43.27 & 51.39 & 50.00 & 19.97 & 29.41 & 8 & 0.69M \\
ORF & 48.90 & 54.43 & 55.40 & 20.95 & 31.99 & -- & -- \\
\bottomrule
\end{tabular}
\caption{Full results overview. Accuracy is reported in percent. Weighted average uses test set sizes: MMLU-R ($n=5632$), ARC-C ($n=1150$), OBQA ($n=500$), and MMLU-Pro ($n=12032$). Size reports bf16 deployment footprint of the projector only. Significance markers are vs.\ C2C under the one-sided exact paired McNemar test: $^{*}p<0.05$, $^{**}p<0.01$, $^{***}p<0.001$. Bold denotes the best fuser result per column.}
\label{tab:app_full_overview}
\end{table}

$^{\dagger}$T2T numbers are from \citet{fu2026cache}. We do not re-evaluate them. C2C does not report T2T on MMLU-Pro. No weighted average is computed.

$^{\ddagger}$LCF-128-9L is the 9-layer critical pruned variant. It keeps 9 of 28 adapter layers. Total trainable parameters: 6.24M. This is a 76.7$\times$ reduction versus C2C's 477.8M fuser.

All rows except T2T use our logits-based evaluation pipeline. The same prompt template and alignment strategy are used. Significance testing is by per-question join. Receiver and sharer are bare frozen models. C2C uses the published \texttt{nics-efc/C2C\_Fuser} checkpoint. LCF rows use our best checkpoints from bottleneck and pruning sweeps.

\section{Training Details and C2C Comparison Protocol}
\label{app:training}

To isolate the effect of the fuser architecture, we match C2C on the frozen model pair, training objective, OpenHermes data, and MCQ evaluation. C2C uses the published full-schedule \texttt{nics-efc/C2C\_Fuser} checkpoint. LCF is trained under the matched protocol below. The main text reports the corresponding wall-clock and per-step training costs.

Across LCF variants, the only architectural variable is latent width; this width also sets the 4$\times$ intermediate MLP dimension shown in Table~\ref{tab:app_hparams_dims}. All variants use the same data, optimizer, schedule, hardware, and seed. We adjust per-device batch size and gradient accumulation to keep VRAM under 80GB on a single A100, while keeping effective batch size at $260 \pm 4$.

\subsection{Matched Training Protocol}

\begin{table}[htbp]
\centering
\footnotesize
\setlength{\tabcolsep}{3pt}
\begin{tabular}{lcccc}
\toprule
Setting & LCF-64 & LCF-128 & LCF-256 & LCF-512 \\
\midrule
Latent dim & 64 & 128 & 256 & 512 \\
Intermediate dim & 256 & 512 & 1024 & 2048 \\
MLP layers & 2 & 2 & 2 & 2 \\
Params / layer & 282K & 693K & 1.9M & 5.9M \\
Total trainable & 7.9M & 19.4M & 53.4M & 165.5M \\
Compression vs.\ joint 2304 & 36$\times$ & 18$\times$ & 9$\times$ & 4.5$\times$ \\
Per-device batch & 10 & 10 & 9 & 8 \\
Grad-accum steps & 26 & 26 & 29 & 33 \\
Effective batch & 260 & 260 & 261 & 264 \\
\bottomrule
\end{tabular}
\caption{LCF training settings by bottleneck width. The architectural
variable is latent width and its corresponding $4\times$ intermediate
MLP width. ``Compression vs.\ joint'' is the ratio of the pre-bottleneck
joint KV dimension ($4HD = 2304$) to the latent dim $d$.}
\label{tab:app_hparams_dims}
\end{table}

\begin{table}[htbp]
\centering
\footnotesize
\setlength{\tabcolsep}{5pt}
\begin{tabular}{ll}
\toprule
Setting & Value \\
\midrule
Receiver / sharer & Qwen3-0.6B / Qwen2.5-0.5B-Instruct \\
Trainable modules & LCF projectors only; both LLMs frozen \\
Dataset & OpenHermes 2.5, 500K-sample subset \\
Train split & 0.99 \\
Max sequence length & 2048 tokens \\
Optimizer & AdamW \\
Learning rate & $1\times10^{-4}$ \\
Weight decay & 0.01 \\
Scheduler & Linear, warmup ratio 0.1 \\
Max grad norm & 1.0 \\
Max steps & 300 \\
Eval / save interval & 50 steps \\
Seed & 42 \\
Hardware & 1$\times$ A100 80GB \\
Wall-clock & $\sim$4--5 h per run \\
Alignment & \texttt{last\_aligned}; training uses \texttt{first}, eval uses \texttt{longest} \\
Gate temperature & 1.0 $\to$ 0.001 over 400 steps; hard at eval \\
Dropout & 0.1 \\
\bottomrule
\end{tabular}
\caption{Settings held constant across all LCF runs.}
\label{tab:app_hparams_constant}
\end{table}

\paragraph{Training cost.}
Under this matched pipeline, LCF-128 takes roughly 52 seconds per optimizer step on a single Colab A100, with an effective batch size of 260. C2C takes roughly 66 seconds per step at batch size 256. The modest wall-clock difference reflects that both methods spend most training time in frozen sharer and receiver forward passes. LCF's size reduction appears primarily as lower trainable-parameter and adapter-memory cost, not as a proportional wall-clock speedup.

\paragraph{Why 300 steps.}
C2C trains for one full OpenHermes epoch (1929 optimizer steps). In both the published C2C curve and our re-run, best MCQ accuracy is reached by step 300. Later checkpoints do not help and can degrade results. LCF shows the same pattern: all bottleneck widths peak between steps 200 and 300. We therefore train LCF ablations for 300 steps. The C2C baseline in our tables is not a 300-step re-run, but the published full-schedule checkpoint. This makes the comparison conservative: we compare 300-step LCF against full-schedule C2C.

\subsection{Eval-Loss Convergence}

Held-out OpenHermes loss converges quickly for all bottleneck widths. Table~\ref{tab:app_eval_loss_dims} reports next-token prediction loss on the held-out 1\% split every 50 steps.

\begin{table}[htbp]
\centering
\footnotesize
\setlength{\tabcolsep}{6pt}
\begin{tabular}{lcccc}
\toprule
Step & LCF-64 & LCF-128 & LCF-256 & LCF-512 \\
\midrule
50  & 2.801 & 1.717 & 3.669 & 2.054 \\
100 & 1.079 & 0.931 & 0.962 & 0.882 \\
150 & 0.895 & 0.892 & 0.902 & 0.875 \\
200 & 0.887 & 0.886 & 0.883 & 0.869 \\
250 & 0.884 & 0.882 & 0.877 & 0.867 \\
300 & 0.880 & 0.878 & 0.874 & 0.860 \\
\bottomrule
\end{tabular}
\caption{Held-out OpenHermes NTP loss across LCF bottleneck widths.}
\label{tab:app_eval_loss_dims}
\end{table}

LCF-64 through LCF-256 converge to within 0.006 nats of each other. This holds despite a 7$\times$ difference in trainable parameters. LCF-512 reaches the lowest NTP loss, but this does not give the best downstream accuracy. It underperforms LCF-256 on three of four benchmarks. Validation NTP loss is not a reliable proxy for fuser quality here. Bottleneck selection requires direct downstream evaluation.

\subsection{Training-Curve Summary}

For all bottleneck widths, train and eval loss track closely. There is no persistent train/eval gap. Larger bottlenecks converge faster at first, but downstream accuracy does not monotonically follow NTP loss. This supports the main ablation result: for this model pair and protocol, useful latent communication peaks at $d=256$. More projector capacity does not always help.

\section{Statistical Significance}
\label{app:significance}

This appendix reports every significance test in the paper. We use a one-sided exact paired McNemar test on per-question correctness, with alternative $H_1:\mathrm{LCF}>\mathrm{C2C}$. Outputs are paired by joining method CSVs on \texttt{(subject, question\_id, true\_answer)}.

For each LCF variant and benchmark, let $b$ count questions where LCF is correct and C2C is wrong, and $c$ count the reverse. Concordant pairs are uninformative. Conditional on $n=b+c$ under $H_0$, $b \sim \mathrm{Binomial}(n,0.5)$, giving the one-sided exact p-value
\[
p = \Pr[\mathrm{Binomial}(n,0.5) \geq b].
\]
We use the exact binomial test rather than the $\chi^2$ approximation because some discordant counts are small. Significance markers are uncorrected: $^{*}p<0.05$, $^{**}p<0.01$, $^{***}p<0.001$.
\subsection{Main LCF Variants}

\begin{table}[htbp]
\centering
\footnotesize
\setlength{\tabcolsep}{4pt}
\begin{tabular}{lcccc}
\toprule
Method & MMLU-R & ARC-C & OBQA & MMLU-Pro \\
\midrule
Receiver  & 34.91 & 39.91 & 39.20 & 16.17 \\
Sharer    & 38.65 & 41.30 & 44.80 & 15.09 \\
C2C       & 42.54 & 54.00 & 50.80 & 19.57 \\
LCF-64   & 38.32 & 46.17 & 43.00 & 17.14 \\
LCF-128  & 45.17$^{***}$ & 56.09 & 52.60 & 22.57$^{***}$ \\
LCF-256  & \textbf{47.53}$^{***}$ & \textbf{57.65}$^{**}$ & 51.80 & \textbf{22.87}$^{***}$ \\
LCF-512  & 45.47$^{***}$ & 56.61$^{*}$ & \textbf{54.40}$^{*}$ & 20.98$^{***}$ \\
\bottomrule
\end{tabular}
\caption{Accuracy (\%) and one-sided exact paired McNemar significance vs.\ C2C. Test sizes: MMLU-R $n=5632$, ARC-C $n=1150$, OBQA $n=500$, MMLU-Pro $n=12032$. Markers are uncorrected.}
\label{tab:app_sig_main}
\end{table}

The strongest pattern appears on the two largest benchmarks. All LCF variants with $d\geq128$ beat C2C on MMLU-Redux and MMLU-Pro at $p<0.001$. On ARC-C and OBQA, test sets are smaller. Gains of 1 or 2 points remain positive but are often not significant. LCF-256's ARC-C gain clears $p<0.01$. LCF-512's OBQA gain clears $p<0.05$.

\subsection{Pruned LCF-128 Variants}

\begin{table}[htbp]
\centering
\footnotesize
\setlength{\tabcolsep}{4pt}
\begin{tabular}{lcccccc}
\toprule
Config & Layers & Params & MMLU-R & ARC-C & OBQA & MMLU-Pro \\
\midrule
Pruned & 19 & 13.17M & $<0.001^{***}$ & 0.09 & 0.20 & $<0.001^{***}$ \\
Critical & 9 & 6.24M & $<0.001^{***}$ & 0.33 & 0.42 & $<0.001^{***}$ \\
\bottomrule
\end{tabular}
\caption{One-sided exact paired McNemar p-values for pruned LCF-128 variants vs.\ C2C.}
\label{tab:app_sig_pruned}
\end{table}

The 19-layer and 9-layer pruned variants remain significant on MMLU-Redux and MMLU-Pro, while ARC-C and OBQA do not clear $p<0.05$. The 6-layer variant does not significantly outperform C2C.

All p-values are computed from paired per-question CSV outputs joined on \texttt{(subject, question\_id, true\_answer)} using \texttt{scipy.stats.binomtest} with \texttt{alternative="greater"}.

\section{Layer Pruning}
\label{app:pruning}

LCF uses one independent projector per receiver layer. For Qwen3-0.6B, this means 28 projectors. Each contributes a residual KV edit at a specific receiver layer. This design makes LCF amenable to post-hoc layer pruning. A trained subset of projectors can be retained at evaluation time without retraining. Dropped projectors simply produce no edit at their layer.

Unlike conventional weight pruning, LCF pruning is structural. Entire layer-projectors are removed, not individual weights. Because projectors are independent across layers, dropping one is equivalent to setting its gate-times-residual contribution to zero. The remaining projectors operate exactly as they did during training.

\paragraph{Parameter accounting.}
LCF-128 trains 28 projectors of about 693K parameters each, for a total of 19.4M trainable parameters. C2C trains 28 fuser layers totaling 477.8M parameters for the same model pair. Keeping $K$ LCF-128 projectors gives roughly $K \cdot 693\mathrm{K}$ trainable parameters. At $K=9$, this is 6.24M, or $76.7\times$ fewer than C2C.

\paragraph{Pruning method.}
Layers are selected in three passes. First, a gate audit removes layers with non-positive trained Gumbel-sigmoid gate logits for both K and V, as these contribute zero signal under the hard evaluation gate. Second, layers that are net harmful by single-layer ablation are removed. Third, the remaining layers are ranked by ablation importance, and the top-$K$ layers are kept.

On the trained LCF-128 checkpoint, six layers are dead by the gate audit:
\[
\{0,1,2,3,5,12\}.
\]
Three more are harmful by ablation:
\[
\{20,24,27\}.
\]
The top-9 retained layers are:
\[
\{6,8,9,10,15,16,17,19,21\}.
\]
No additional fine-tuning is done after pruning.

\begin{table}[!htbp]
\centering
\scriptsize
\setlength{\tabcolsep}{2pt}
\caption{LCF-128 layer pruning. $K$ is the number of retained projectors. Parameter counts are projector-only.}
\label{tab:app_pruning_sweep}
\begin{tabular}{lrrrrrrrr}
\toprule
Configuration & $K$ & Size (MB) & Params & vs.\ C2C & MMLU-R & ARC-C & OBQA & MMLU-Pro \\
\midrule
C2C & 28 & 956 & 477.85M & 1.0$\times$ & 42.54 & 54.00 & 50.80 & 19.57 \\
\midrule
LCF-128 full & 28 & 39 & 19.39M & 24.5$\times$ & 45.17 & 56.09 & 52.60 & 22.57 \\
Alive only & 22 & 31 & 15.25M & 31.3$\times$ & 45.12 & 55.91 & 53.40 & 22.40 \\
Drop dead + harmful & 19 & 26 & 13.16M & 36.2$\times$ & 45.90 & 55.83 & 52.60 & 22.23 \\
Top-9 critical & 9 & 13 & 6.23M & 76.3$\times$ & 45.65 & 54.70 & 51.40 & 21.45 \\
Top-6 critical & 6 & 8 & 4.16M & 114.5$\times$ & 43.27 & 51.39 & 50.00 & 19.97 \\
\bottomrule
\end{tabular}
\end{table}

The top-9 configuration is the main pruned result. With only 6.24M trainable parameters, it remains above C2C on all four benchmarks: $+3.11$ on MMLU-Redux, $+0.70$ on ARC-C, $+0.60$ on OBQA, and $+1.88$ on MMLU-Pro. The gains on MMLU-Redux and MMLU-Pro are statistically significant under the one-sided exact McNemar test in Appendix~\ref{app:significance}.
The top-9 configuration is the main pruned result. With only 6.24M trainable parameters, it remains above C2C on all four benchmarks: $+3.11$ on MMLU-Redux, $+0.70$ on ARC-C, $+0.60$ on OBQA, and $+1.88$ on MMLU-Pro. The gains on MMLU-Redux and MMLU-Pro are statistically significant under the one-sided exact McNemar test in Appendix~\ref{app:significance}.

\paragraph{Why pruning works.}
The trained gates already disable several layers, and ablations show some active layers are net harmful. Edits at one receiver layer propagate through later layers via the residual stream, so adjacent edits can have correlated effects. The pruning sweep suggests most useful LCF-128 signal is concentrated in a mid-to-deep subset of receiver layers.

\paragraph{Limitations.}
The retained layer indices are specific to the trained Qwen2.5-0.5B-Instruct $\to$ Qwen3-0.6B checkpoint. The pruning method generalizes, but the selected layers may differ for other model pairs, training runs, or tasks.

\section{LCF-X Training Details}
\label{app:lcfx_training_details}

LCF-X is trained on HotpotQA-bridge with the same joint-KV bottleneck projector as LCF-128, but includes two changes: a hierarchical attention pool (see Section~\ref{sec:lcfx}) and a gate-logit initialization of $+1.0$. Both Qwen3-0.6B models remain frozen. Only the LCF projector and the two pool queries per layer, $q_{\text{base}}^{(\ell)}$ and $q_{\text{layer}}^{(\ell)}$, are updated.

The agent setting differs from the base LCF in two ways. First, we use a symmetric Qwen3-0.6B/Qwen3-0.6B pair because both models must process partial context independently. Second, training is task-specific: LCF-X is trained directly on HotpotQA-bridge, so the latent channel learns to route partitioned multi-hop evidence.

\subsection{Hyperparameters}

\begin{table}
\centering
\footnotesize
\setlength{\tabcolsep}{6pt}
\begin{tabular}{lc}
\toprule
Setting & Value \\
\midrule
Receiver / sharer & Qwen3-0.6B / Qwen3-0.6B \\
Trainable parameters & $\approx$25.8M projector + 57K pool \\
Pool variant & hierarchical, shared base query per layer \\
Pool query init & Kaiming $\times 0.02$ \\
Gate logit init & $+1.0$ \\
Up-proj Kaiming scale & 0.06 \\
Latent dim & 128 \\
Intermediate dim & 512 \\
MLP layers & 2 \\
Dataset & HotpotQA-distractor train, bridge filter \\
Dev holdout & 2.8\% \\
Train items & $\sim$70.7K \\
Max sequence length & 2{,}048 tokens \\
Optimiser & AdamW \\
Learning rate & $1\times10^{-4}$ \\
Warmup & 200 steps, linear \\
Per-device batch & 16 \\
Grad accum & 16 \\
Effective batch & 256 \\
Epochs & 1 \\
Optimiser steps & $\sim$277 \\
Gate temperature anneal & $1.0 \to 0.5$ \\
Eval / save interval & 50 steps \\
Seed & 0 \\
Wall-clock & $\sim$50 min \\
\bottomrule
\end{tabular}
\caption{LCF-X training hyperparameters.}
\label{tab:app_lcfx_hparams}
\end{table}

Training partitions the sharer's input using HotpotQA's five-paragraph structure, so $P=5$ during training. The architecture is partition-agnostic: $P$ is not a parameter shape and need not remain fixed at inference; see Section~\ref{app:lcfx_span_invariance}.

We initialize the gate logit to $+1.0$ instead of $0.0$. In early LCF-X runs, zero-initialized gates stayed near sigmoid$(0)=0.5$ because the residual branch was small, producing weak gate gradients. Setting it to $+1.0$ gives the residual branch a meaningful gain at step 0, sigmoid$(1.0)=0.73$, while preserving trainability.

\subsection{Training Trajectory}

We log dev F1/EM every 50 steps during training. Validation metrics are reported on the held-out HotpotQA-bridge split ($n{=}5{,}899$) at the final checkpoint.

\begin{table}
\centering
\footnotesize
\setlength{\tabcolsep}{6pt}
\begin{tabular}{lcc}
\toprule
Step & Dev F1 & Dev EM \\
\midrule
50  & 26.85 & 18.80 \\
100 & 37.82 & 29.80 \\
150 & 40.61 & 33.50 \\
200 & 39.79 & 31.75 \\
250 & 41.38 & 34.85 \\
Final (276) & \textbf{41.45} & \textbf{34.85} \\
\bottomrule
\end{tabular}
\caption{LCF-X training trajectory on the 2{,}000-item development holdout. Final evaluation is run once on HotpotQA-bridge val using the final checkpoint.}
\label{tab:app_lcfx_training_curve}
\end{table}

After warmup, performance improves steadily, with the best checkpoint at the end of the single-epoch run. The span-invariance check in Section~\ref{app:lcfx_span_invariance} rules out dependence on the specific five-paragraph partition used during training.

We use the final checkpoint for evaluation. On HotpotQA-bridge val ($n{=}5{,}899$), the final checkpoint achieves 35.13 F1 and 25.28 EM.

\subsection{Architectural Cost}

The LCF-X pool adds two query vectors per receiver layer: a base query $q_{\text{base}}^{(\ell)} \in \mathbb{R}^{H \times d}$ for the first pooling stage and a layer query $q_{\text{layer}}^{(\ell)} \in \mathbb{R}^{H \times d}$ for the second pooling stage:
\[
|q_{\text{base}}^{(\ell)}| + |q_{\text{layer}}^{(\ell)}| = 2 H d.
\]
For Qwen3-0.6B, with $H{=}8$ KV heads and $d{=}128$,
\[
2 \cdot 8 \cdot 128 = 2{,}048
\]
parameters per layer, totaling $57{,}344$ across the 28-layer receiver. This is $0.22\%$ of the LCF-X projector's $25.87$M trainable parameters and about $0.005\%$ of the frozen two-model system.

Critically, the pool size is independent of $P$: the same two queries are reused for every source span, regardless of how many spans the deployment chooses. The pool, therefore, supports variable span counts without retraining or changing parameter shapes.

\subsection{Span-Scheme Invariance}
\label{app:lcfx_span_invariance}

LCF-X is trained with the natural HotpotQA five-paragraph partition. Because the pool has no \(P\)-dependent parameters, the same trained checkpoint admits different partition schemes at inference. We test this directly by holding the trained weights fixed and evaluating on HotpotQA-bridge validation under paragraph-, token-, and sentence-level span schemes.

\begin{table}[!htbp]
\centering
\footnotesize
\setlength{\tabcolsep}{4pt}
\begin{tabular}{lcccccc}
\toprule
Span scheme & $P$ & F1 & EM & $\Delta$F1 & $\Delta$EM & \\
\midrule
Natural paragraphs              & 5.0  & 35.13 & 25.28 & ---      & ---  \\
Paragraph halves                & 10.0 & 35.21 & 25.48 & \(+0.09\) & \(+0.20\) \\
Token windows, no overlap       & 4.4  & \textbf{35.45} & \textbf{25.50} & \(\mathbf{+0.32}\) & \(\mathbf{+0.22}\) \\
Token windows, 50-token overlap & 5.8  & 35.37 & 25.36 & \(+0.25\) & \(+0.08\)  \\
Sentence spans                  & 23.4 & 35.17 & 25.31 & \(+0.05\) & \(+0.03\) \\
\bottomrule
\end{tabular}%
\caption{Inference-time span-scheme ablations on a single trained LCF-X checkpoint. The span scheme varies across rows; trained weights are identical.}
\label{tab:app_lcfx_span_ablation}
\end{table}

F1 stays within a \(0.32\)-point band of the natural-paragraph reference, even across a \(4.7\times\) span-count range and three granularities. The best result comes from 200-token windows with no overlap, improving F1 by \(+0.32\) and EM by \(+0.22\). All schemes use the same 55.1 GB VRAM footprint, indicating that changing the span partition changes the source summary structure without changing the deployed checkpoint.

At inference time, \(P\) can be chosen to match the input structure, using retrieved passages, sentence chunks, sliding windows, or a single span. No retraining or deployment-specific fine-tuning is needed.

\end{document}